\documentclass{article}

\usepackage{arxiv}

\usepackage[utf8]{inputenc} 
\usepackage[T1]{fontenc}    
\usepackage[unicode=true,pagebackref=true,
colorlinks,linkcolor=blue,citecolor=red,final]{hyperref}       
\usepackage{url}            
\usepackage{booktabs}       
\usepackage{amsfonts}       
\usepackage{nicefrac}       
\usepackage{microtype}      
\usepackage{lipsum}
\usepackage{graphicx}
\usepackage[export]{adjustbox}
\usepackage[skip=-0.3pt,font=tiny]{subcaption}
\usepackage{multirow}
\graphicspath{ {./images/} }

\title{Denoising single images by feature ensemble revisited}

\author{
 Masud An Nur Islam Fahim \\
  Chosun University\\
  Gwangju, South Korea \\
  \texttt{mostofafahim21@gmail.com} \\
   \And
 Nazmus Saqib \\
  Chosun University\\
  Gwangju, South Korea \\
  \texttt{nsaqib1995@gmail.com} \\
  \And
 Shafkat Khan Siam \\
 Chosun University\\
  Gwangju, South Korea \\
  \texttt{shafkat.kh022@gmail.com} \\
  \AND
  Ho Yub Jung\\
  Chosun University\\
  Gwangju, South Korea \\
  \texttt{jung.ho.yub@gmail.com}\\
}

\begin{document}
\maketitle
\begin{abstract}
Image denoising is still a challenging issue in many computer vision sub-domains. Recent studies show that significant improvements are made possible in a supervised setting. However, few challenges, such as spatial fidelity and cartoon-like smoothing remain unresolved or decisively overlooked. Our study proposes a simple yet efficient architecture for the denoising problem that addresses the aforementioned issues. The proposed architecture revisits the concept of modular concatenation instead of long and deeper cascaded connections, to recover a cleaner approximation of the given image. We find that different modules can capture versatile representations, and concatenated representation creates a richer subspace for low-level image restoration. The proposed architecture's number of parameters remains smaller than the number for most of the previous networks and still achieves significant improvements over the current state-of-the-art networks.  
\end{abstract}


\section{Introduction}
Image denoising is a classic problem in the low-level vision domain. For a given image $\mathcal{X}$, it goes through the following mapping to create its noisy counterpart.

\begin{equation}
    \mathcal{Y} = \mathcal{X} + \mathcal{N} 
\end{equation}

Here, $\mathcal{Y}$ is the noisy observation, where $\mathcal{N}$ is the additive noise on a clean image $\mathcal{X}$. Denoising is an ill-posed problem, with no direct means to separate the source image and corresponding noise. Hence, researchers follow the best possible approximation of $\mathcal{X}$ from $\mathcal{Y}$ with corresponding algorithmic strategies. 

Typical methods without machine learning involve employing efficient filtering techniques such as NLM\cite{NLM}, BM3D\cite{BM3D}, median\cite{median}, Weiner\cite{weiner}, etc. Due to their limited generalization capability, additional knowledge-based priors or matrix properties were integrated into the denoising strategies. However, despite certain improvements with prior-based methods, many concerns remain unresolved, such as holistic fidelity or the choice of priors.

\begin{figure}
\centering
\begin{subfigure}[l]{0.18\textwidth}
      \includegraphics[width=18mm,height=28mm]{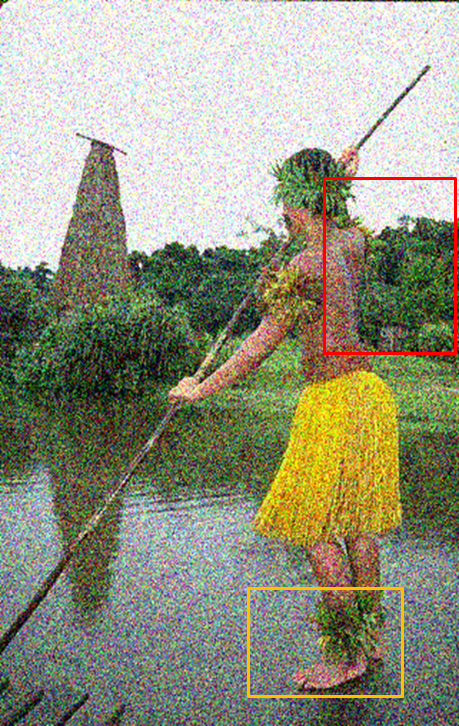}
      \centering\caption{\tiny \textbf{Noisy}}
\end{subfigure}
\begin{subfigure}[r]{0.35\textwidth}
    \centering
    \begin{subfigure}{0.32\textwidth}
    \includegraphics[width=\textwidth,height=14mm]{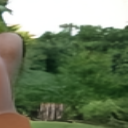}  
    \end{subfigure}
    \begin{subfigure}{0.32\textwidth}
    \includegraphics[width=\textwidth,height=14mm]{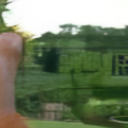}
    \end{subfigure}
    \begin{subfigure}{0.32\textwidth}
    \includegraphics[width=\textwidth,height=14mm]{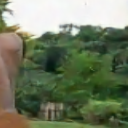}  
    \end{subfigure}\\
    \begin{subfigure}{0.32\textwidth}
    \includegraphics[width=\textwidth,height=14mm]{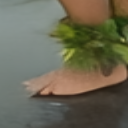}  
    \caption{\tiny \textbf{Ground Truth}}
    \end{subfigure}
    \begin{subfigure}{0.32\textwidth}
    \includegraphics[width=\textwidth,height=14mm]{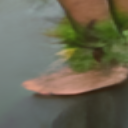}
    \caption{\tiny \textbf{DEAMNet}\cite{deamnet}}
    \end{subfigure}
    \begin{subfigure}{0.32\textwidth}
    \includegraphics[width=\textwidth,height=14mm]{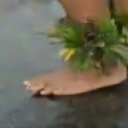} 
    \caption{\tiny \textbf{Proposed}}
    \end{subfigure}
\end{subfigure}\\
    \caption{ Demonstration of our contribution. An image from the BSD68 dataset with AWGN noise $\sigma$ = 50. Here, the first column shows the ground truth, followed by the inference from the DEAMNet\cite{deamnet} and the final column shows the proposed result. From a side-by-side comparison, our reconstruction restores a closer approximation of the ground truth than the referred study with finer details within both foreground and background patches.}
    \label{fig:my_label}
\end{figure}

Convolution neural network (CNN) denoising methods later offer an unprecedented improvement over the previous strategies through their customized learning setup. Usually, CNN methods offer better performance through brute force learning \cite{RIDnet}, tricky training strategy \cite{NVNet}, or inverting image properties \cite{GradNet} by various proposals. We observed gradual improvements over the years for denoising solutions. However, these methods with pure brute force mapping sometimes face fidelity issues within challenging noisy images. Furthermore, due to the lack of generalization properties, the methods provide reconstructed images that often result in cartoonized smoothing. 

In contrast, the proposed approach rebuilds a previous ensemble-oriented denoising network that can successfully estimate cleaner image with less cartoon-like smoothing. For the design of the proposed denoising network, we carefully maximized detail restoration by providing a variety of low-level ensemble features while keeping the network relatively shallow to prevent an oversized receptive field. In summary, our study has the following contributions:
\begin{itemize}
    \item We propose a shallow ensemble approach through feature concatenation to create a large array of feature combinations for low-level image recovery.
    \item Due to the ensemble of multiple modules, our model successfully returns fine details compared to previous data-driven studies.
    \item The parameter space is relatively small compared to the contemporary methods with computationally fast inference time.
    \item Finally, the proposed study shows better performance with a different range of synthetic noise and real noise without the cartoonization effect. See fig. \ref{fig:my_label}.
\end{itemize}

\section{Related Work}
\subsection{Filtering based schemes}

Traditional filtering approaches aim for a handcrafted filters for noise and image separation. These studies \cite{weiner,median} utilize low-pass filtering methods to extract the clean images from the noisy images. The iterative filtering approach adopts progressive reduction for image restoration\cite{knaus}. Additionally, several methods use nonlocal similar patches for noise reduction based on the similarity between the counterpart patches in the same image. For example, NLM\cite{NLM} and BM3D\cite{BM3D} assume redundancy within patches from a given image for noise reduction. Nonetheless, these methods usually produce flat approximations, as the given image severely degrades the noisy image quality with a heavy noise presence.

\subsection{Prior based schemes}
Another group of studies focuses on selecting priors for the model that produce clean images when optimized. These methods reformulate the denoising problem as a maximum a posteriori (MAP)-based optimization problem, where the image prior regulates the performance of the objective function. For example, the studies \cite{42,14} assume sparsity as the prior for their optimization process. The primary intuition is to represent each patch separately through the help of a function. Xu \textit{et al.} \cite{WNNM} performed real-world image denoising by proposing a trilateral weighted sparse coding scheme. Other studies \cite{WNNM,55,54,57} focus on rank properties to minimize their objective function. Weighted nuclear norm minimization (WNNM) \cite{WNNM} calculates the nuclear norm through a low-rank matrix approximation for image denoising. Additionally, there are several complex model-based derivations using graph-based regularizers for noise reduction. However, their performance degrades monotononically for noisier areas, and recovering the detailed information is sometime difficult \cite{37,L0}.  Additionally, these methods generally output significantly varying results depending on their prior parameters and the respective target noise levels. 

\subsection{Learning based schemes}
Due to the availability of paired data and the current success of CNN modules, data-driven schemes have achieved significant improvement in separating clean images from noisy images. Recent CNN studies, such as DnCNN\cite{DnCNN} and IrCNN\cite{IrCNN},  utilized the residual connection for the estimating noise removal map before inference. Both of them evaluated the clean image without taking any priors regarding the structure or noise. They achieved enhanced performance by using a noncomplex architecture with repeated convolutional, batch normalization, and ReLU activation function blocks. However, these methods can fail to recover some of the detailed texture structure in the presence of heavy noise area.

Trainable Nonlinear Reaction-Diffusion (TRND) \cite{TRND} uses prior in their neural network and extends the non-linear diffusion algorithm for noise reduction. However, the methods suffer from computational complexity due to requiring a vast number of parameters. Similarly, the nonlocal color net \cite{ridnet39} utilizes the nonlocal similarity priors for the image denoising operation. Although priors mostly aid in denoising, there are some cases where the adaptation of the priors degrades the denoising performance. Very recently, DEAMNet\cite{deamnet} surpasses the previous state-of-the-art results by using adaptive consistency prior.

With the success of the DnCNN\cite{DnCNN}, two similar networks called ``Formatting net" and ``DiffResNet" are proposed with different loss layers \cite{RIDnet}. Later, Bae \textit{et al.} \cite{Baeetal} proposed a residual learning strategy based on the improved performance of a learning algorithm using manifold simplification, providing significantly better performance. After that, Anwar \textit{et al.}\cite{anwar} proposed a cascaded CNN architecture with feature attention using self-ensemble to boost the performance. 

Few recent approaches \cite{NC,ridnet31} follow the blind denoising strategy. CBDNet \cite{ridnet31} proposed a blind denoising network consisting of two submodules: noise estimation and noise removal by incorporating multiple losses. However, their performance may be limited by manual intervention requirements, and a slightly lower performance on real-world noisy images. In comparison, FFDNet \cite{ffdnet} achieves enhanced results by proposing the nonblind Gaussian denoising network. Consequently, RIDNet \cite{RIDnet} utilizes perceptual loss with $\ell_{2}$ apart from the DnCNN architectures for noise removal and achieves significant success by introducing single-stage attention denoising architecture from real and synthetic noises. Recently, Liu \textit{et al.} \cite{GradNet} introduced GradNet by revisiting the image gradient theory of neural networks.

\section{Methodology}
\subsection{Baseline supervised architecture}

\begin{figure}[htbp]
    \centering
    \includegraphics[width=0.8\textwidth, height= 1.6in]{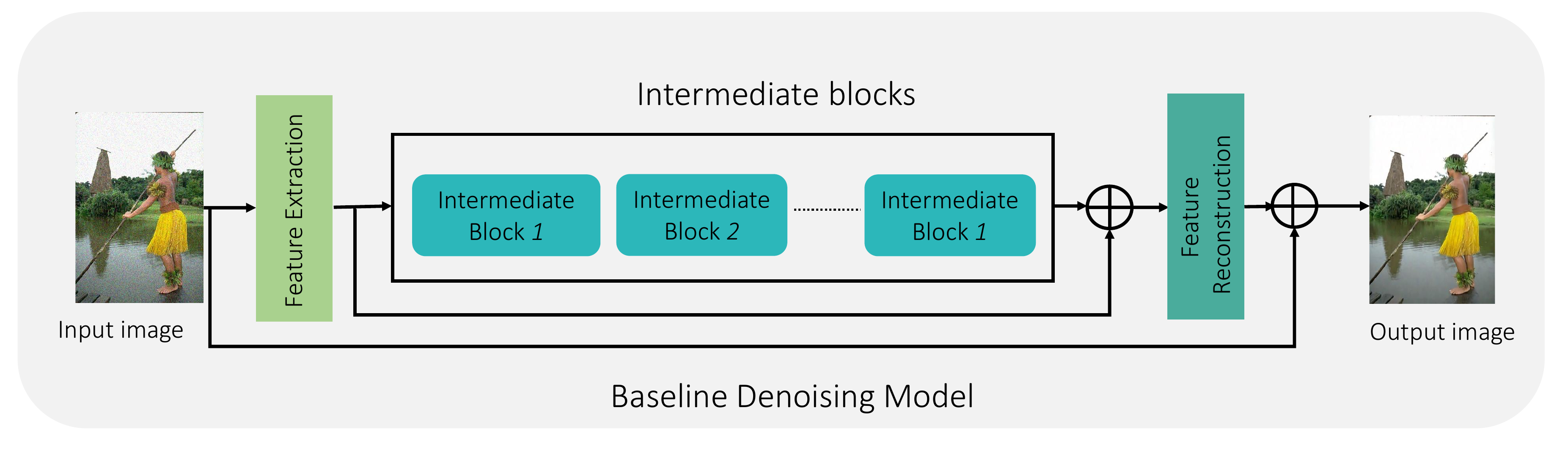}
    \caption{ This figure shows the general baseline architecture for the denoising model, which usually consists of the more prolonged feature extraction phase with cascaded modules, which begin right after the initial feature collection and end with the final residual aggregation. }
    \label{architecture_base}
\end{figure}

Recently, supervised model-based denoising methods embedded a similar baseline formation in their proposals \cite{RIDnet,aindnet}. In brief, it is possible to compartmentalize the baseline architecture in fig. \ref{architecture_base} into three distinct modules: initial feature learner, large intermediate processor, and final reconstruction part. Typically, the primary module consists of a single layer that serves the purpose of the initial feature learner or initial noise estimator. For any noisy image $I_{n}$, the representation of the initial block $E_{p}$ is as follows:
\begin{equation}
     E_{p} =  \beta(I_{n})
\end{equation}
where $\beta$ is the initial convolutional layer for basic feature training. Followed that, we see the main restoration part of the given network through the help of an intermediate processor. Typically, this intermediate layer is a very long cascaded connection of the unique feature extractor units. From time to time, we often observe the presence of long residual-dense or residual-attention blocks as the backbone of such setup.

Now, if the intermediate block is $\mathcal{M}$; the cascaded representation of the intermediate processing stage is as follows:
\begin{equation}
    E_{i} = \mathcal{M}_{j}(\mathcal{M}_{i-1}(.....(E_{p})..))
\end{equation}
where $\mathcal{M}_{i}$ represents the $i^{th}$ instance of the learning stage of the intermediate block and $E_{i}$ is the corresponding outcome of the intermediate layer.


The final reconstruction module operates through a residual connection followed by a consecutive final convolution. If $\mathcal{R}$ is the final reconstruction stage before the output, then the recovered image $I_{r}$ is a combination of $E_{i}$, $E_{p}$, and $I_n$.
\begin{equation}
  I_{r} = \mathcal{R} (E_{i},E_{p}, I_{n}) = f_{N}(I_{n})
\end{equation}
Here, $f_{N}$ denotes the overall neural network, $I_{n}$ is the noisy input image, and $I_{r}$ is the recovered image. A typical choice of the cost function for this task involves $\ell_{1}$ or $\ell_{2}$ loss. There are other customized loss functions available, such as weighted-augmentation of different loss functions that integrate spatial properties or relevant regularization \cite{RIDnet}. In general, the network is optimized by minimizing the difference from clean images.
\begin{equation}
    \zeta(\theta) = \frac{1}{N}\sum_{i=1}^N||f_N(\theta, {I_{n}^i})-I_{c}^i||_1
\end{equation}
Here, $\theta$ is the learnable parameter, $I_{n}^{i}$ is the  noisy image, and $I_{c}^{i}$ is the corresponding clean image. Most of the baseline network parameters are placed in the intermediate learning block.

\subsection{Proposed architecture}
In contrast, we design our network to allocate more resources to the concatenated learned features. Instead of developing a basic learning block for long cascading connections, we go for depth by proposing various individual feature learning blocks. The proposed network is focused on delivering richer and diverse low level features. To further reduce complexity, we avoid using attention operation, which is typically more expensive. More details are provided in fig. \ref{architecture_overview} and the following subsection.

\subsubsection{Initial feature block}
\begin{figure*}
    \centering
    \includegraphics[width=\textwidth, height= 2.7in]{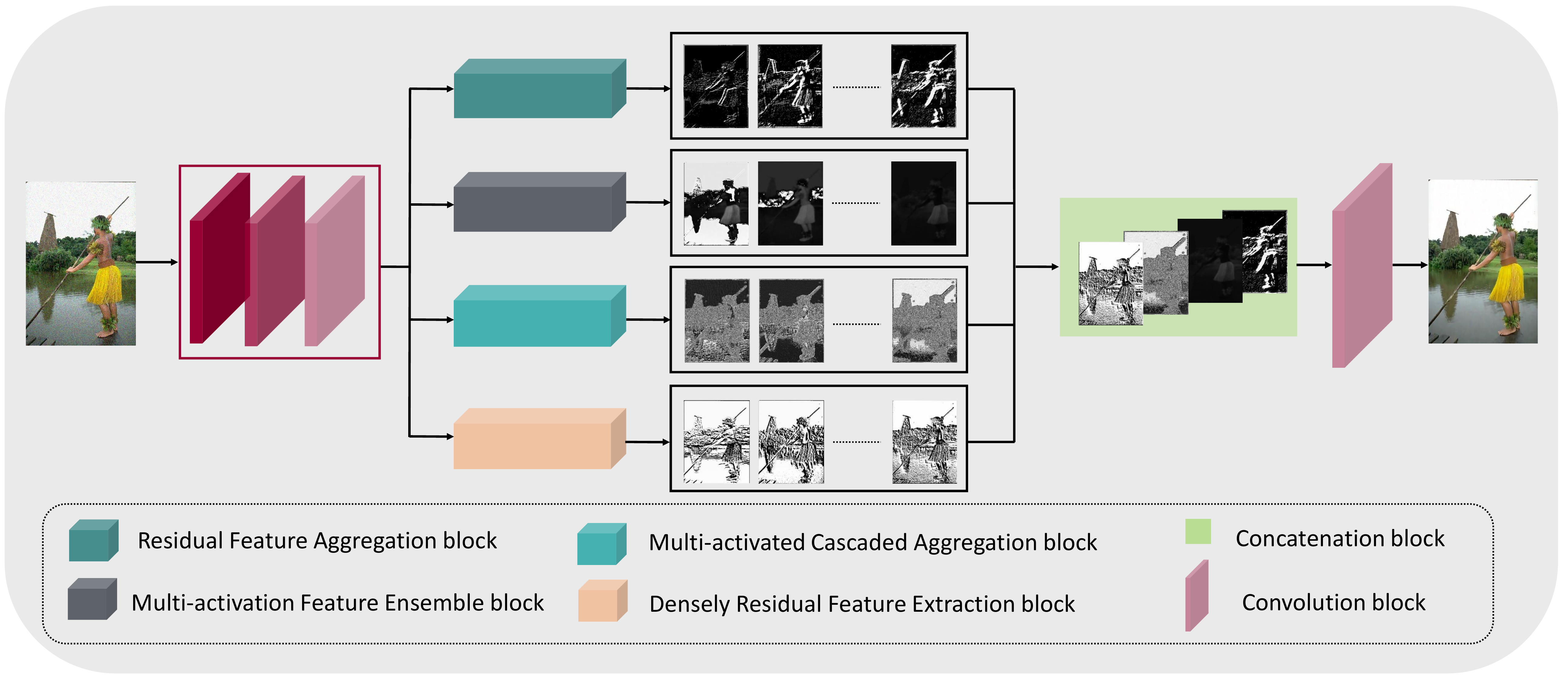}
    \caption{ In the above figure, we present the overall diagram of the proposed architecture for image denoising. Our pipeline first extracts the initial feature using consecutive convolution operation, followed by the four modules for feature refinements. These modules are standing upon the customized convolution and residual setup with supportive activation functions. After refinement, we concatenate all the refined feature maps into a single layer, followed by a final dilated convolution to make the inference.}
    \label{architecture_overview}
\end{figure*}

Three consecutive convolution layers are used to extract initial features for the network. The layers are equal in-depth, but their kernel sizes are in descending order. The input image goes through the $5\times5$ convolution operation at the first layer, followed by $3\times3$ convolution, and ends at $1\times1$ pixelwise operation. A larger kernel size makes use of a larger neighborhood of input features and estimates the representations on a larger receptive fields. By limiting the kernel size and the number of layers, the network learns to focus on the smaller receptive fields and disregards broader view, which we argue to be less meaningful in low-level vision tasks such as denoising. Therefore, the purpose of the primary layer is to project the representation for the denoising features from a smaller receptive field into individual responses which can be further diversified in the next four block modules in fig. \ref{architecture_overview}.

\subsection{Four modules for feature refinement}
Before presenting the four modules for feature refinement, we will cover the convolution, activation functions, and residual connections used in the modules. Even though the attention mechanism is a common choice to learn richer representations, we can still find a similar or better result without it in this study. Additionally, our selection of residual blocks are no longer than six consecutive connections. The fundamental operations for our modules are introduced below.

\textbf{Convolution.} In the internal convolution operation, our choice of kernels varies from $1\times1$ to $7\times7$. Due to such a range, our network is naturally focused on both smaller and larger receptive fields.

\textbf{Activation functions.} Recent advancements in nonlinear activation functions have shown that better performance is achievable through the interconnected operation of different activation representations that are compacted into a single function. Hence, we choose the SWISH\cite{swish} and MISH\cite{mish} activations in addition to the ReLU operations. As a result, our network learns from diverse representations obtained from various parallel activated functions.

\textbf{Residual connections.} It will become redundant to mention the efficacy of the residual connections in the vision tasks. In the literature, we can see that the customization of residual connections varies within the task. In the original ResNet paper \cite{resnet}, the authors included batch-normalization between the convolution layer, followed by the ReLU layer. In our study, we use the convolution layers, which are separated by the ReLU layer. This choice of the ReLU sandwich residual connection is prevalent in regression tasks \cite{edsr}. 

We will focus on the major processing modules below with the description of the utilized blocks. We propose four processing modules that perform the refinement operations on the initial features. The following subsections covers descriptions and the basic reasoning behind the proposed architecture.

\subsubsection{Residual Feature Aggregation Module}
\begin{figure}[htbp]
    \centering
    \includegraphics[width=0.6\textwidth, height=1.5in]{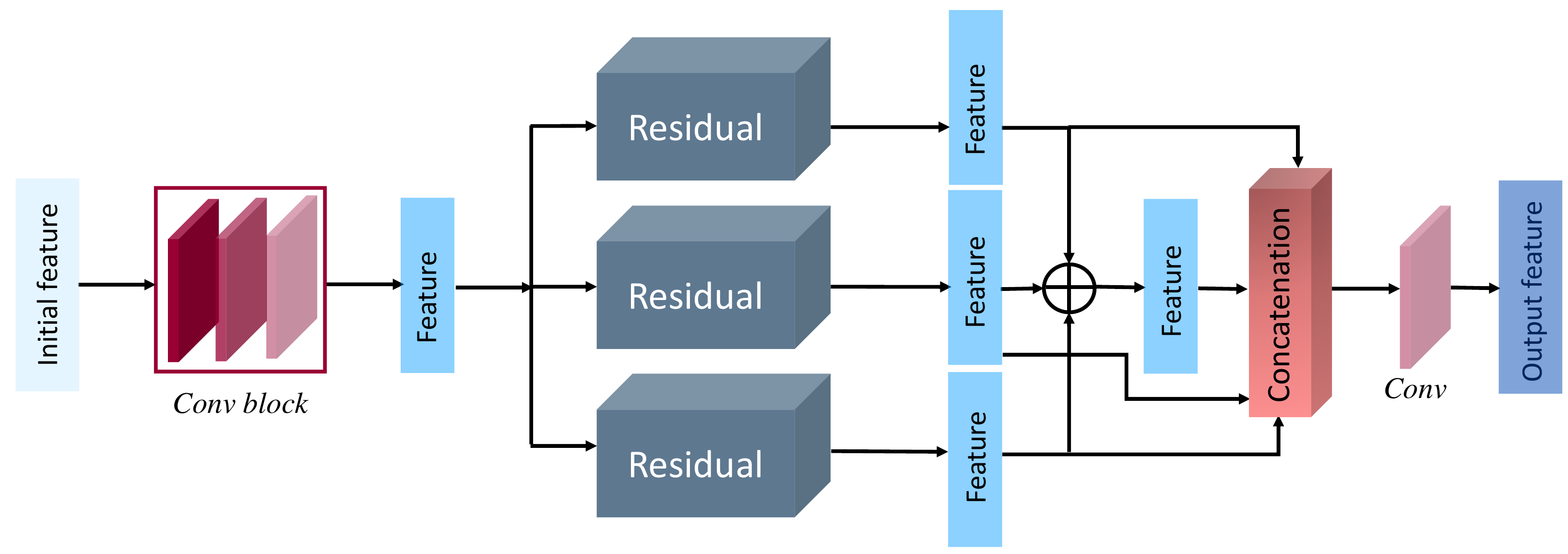}
    \caption{Residual feature Aggregation Module}
    \label{residual}
\end{figure}

In our residual feature aggregation module, we use the aforementioned residual blocks as our underlying design mechanism. In the construction of this module, we took inspiration from the traditional pyramid feature extraction\cite{Pyramid} and aggregation, which has been very influential in computer vision. A typical pyramid setup is motivated by the needs for multiscale feature aggregation, which, in essence, utilizes low-frequency information along with high frequency features. However, subsequent downsampling process is a lossy operation by nature. To mitigate information loss for low-frequency features, we choose to employ the concurrent residual blocks on the same initial features through three different kernel sizes. Naturally, our kernel choice ranges from $1\times1$ to $5\times5$. See fig. \ref{residual}. Hence, a larger kernel allows us to learn the features from larger area of image, while the $1\times1$ kernel operation allows us to maintain the initial receptive field and make use of more high-frequency information. We aggregate the response from all three residual block to learn the overall multiscale impact of the initial features. Finally, a typical $3\times3$ convolution with standard depth gives us the $n$ number of diverse representations from this module. As a result, our model can learn the important multiscale features without going through a pooling operation.

\subsubsection{Multi-Activation Feature Ensemble}
Activation functions are unavoidable components for neural network construction that aid the learning operation by projecting the impactful information to the next layer. Hence, widely different nonlinear functions are available as activation functions in all sorts of neural networks for various utilities. ReLU is the most widely used activation, which at heart is a `positive pass' filter. However, in some cases, zero-out negatives and a discontinuity in the gradient are argued to be unhelpful in the optimization process. To address some of its weaknesses, SWISH\cite{swish} and MISH\cite{mish} were proposed with smooth gradients while maintaining a similar positive pass shape of ReLU. A recent experiment \cite{mish} shows that these activation functions provide a smoother loss landscape than the ReLU.

\begin{figure}[htbp]
    \centering
    \includegraphics[width=0.5\textwidth, height=2in]{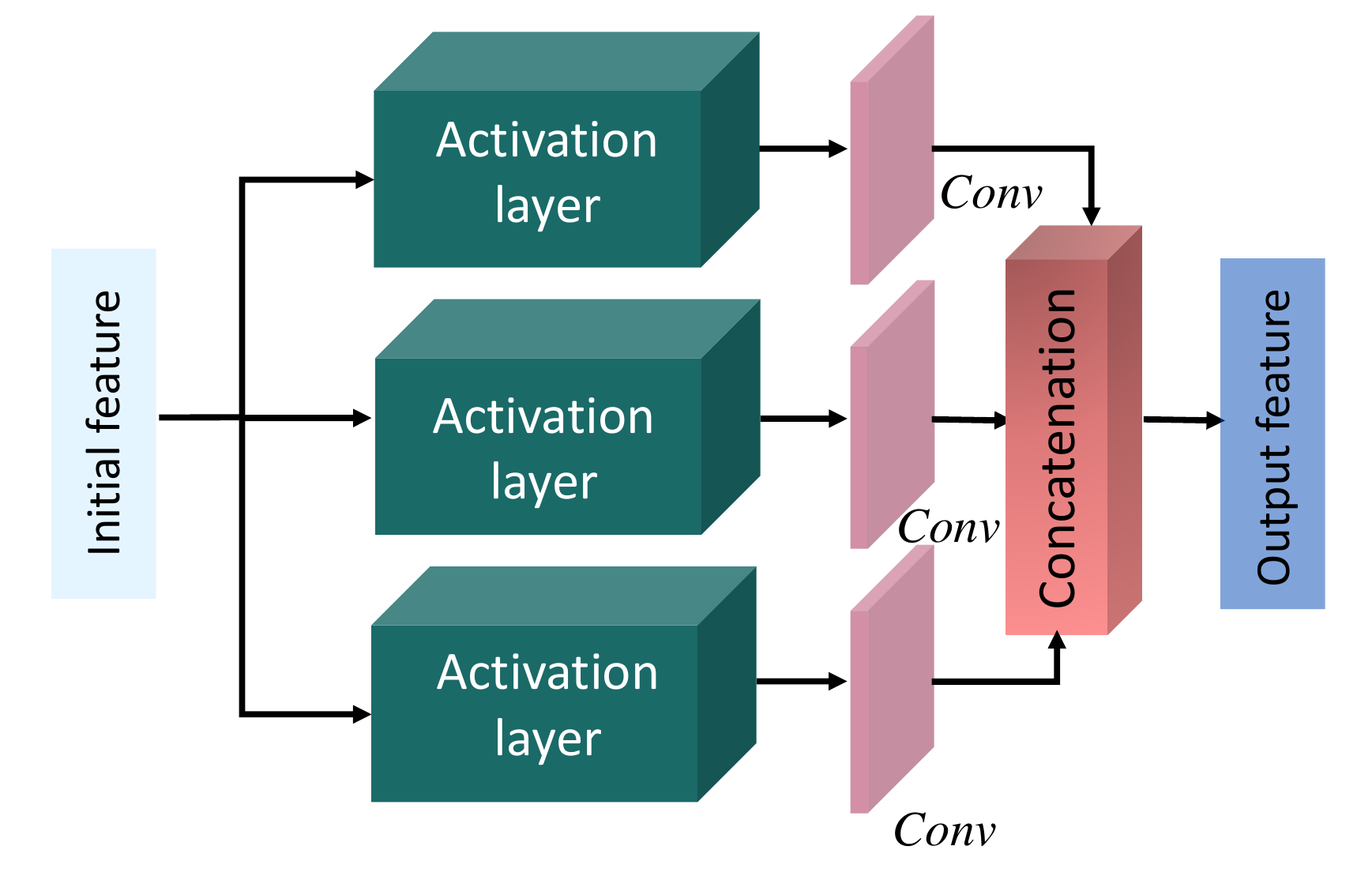}
    \caption{Multi-Activation Feature Ensemble Module}
    \label{actc}
\end{figure}

Nonetheless, we incorporate all three activation functions. See fig. \ref{actc}. SWISH, MISH, and ReLU activation are applied to the initial features, followed by a convolution layer. The subsequent responses are concatenated into a single tensor to learn from the integrated representation of varying activation functions. No further kernels and residual blocks are utilized for this module. The initial feature results of these modules are ensembled with the responses of the other three modules, but the multi-activation are also integrated into the multi-activated cascaded aggregation module described in the next subsection.

\subsubsection{Multi-activated Cascaded Aggregation}
\begin{figure}[htbp]
    \centering
    \includegraphics[width=0.8\textwidth,height=1.8in]{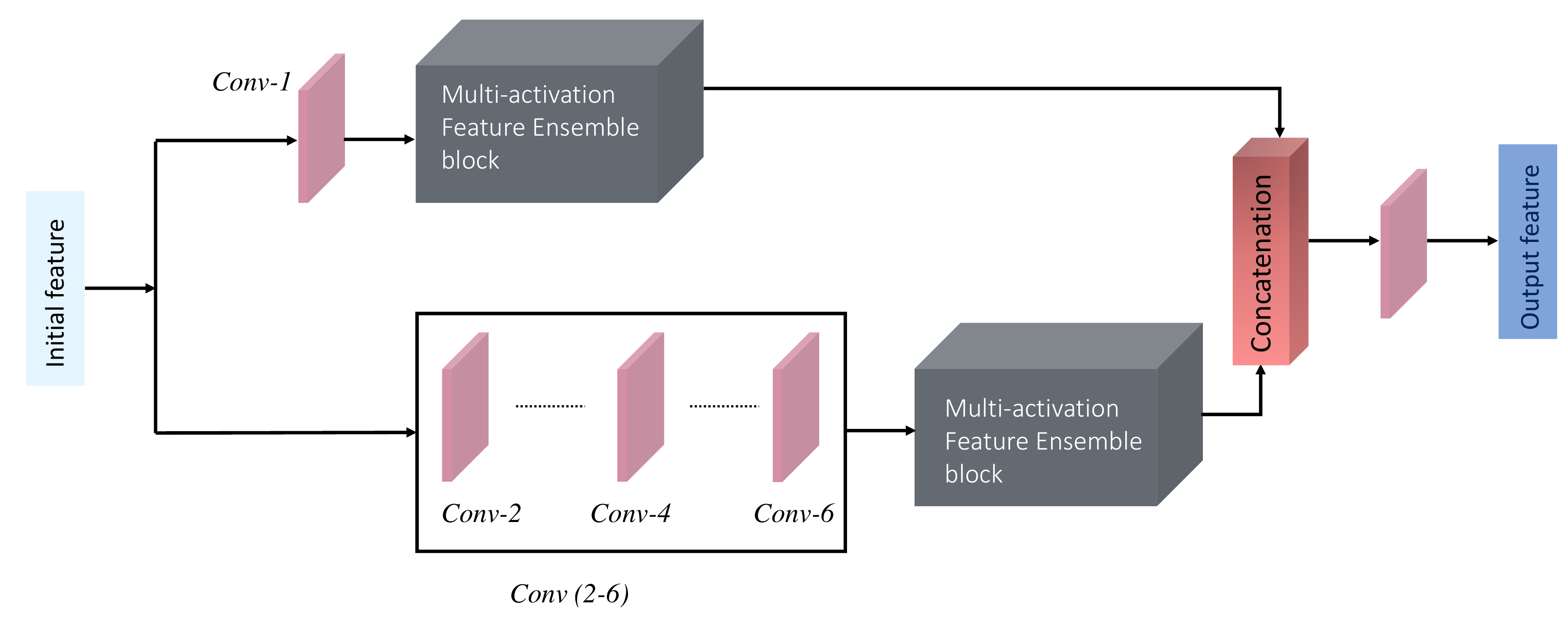}
    \caption{Residual feature Aggregation Module}
    \label{cascaded}
\end{figure}

In this module, both shallow and relatively deeper layer features are concatenated. Typically, a deep consecutive convolution operation is formulated after the initial feature extraction, and the conventional thinking is to build a deeper network for complex problems. However, we add a single convolution layer feature to complement the deeper layer features because we believe that shallower interpretation might be more appropriate for low-level vision problems. See fig. \ref{cascaded}.

For a single convolution path, a $3\times3$ kernel size is chosen with the same depth as the initial features. For the deeper path, five consecutive convolution layers with different kernel sizes were used. The activation functions between the layers are ReLUs, however, for both paths, multi-activation feature ensemble are implemented as described earlier. Both the shallow and deeper responses are concatenated followed by another convolution layer.

\subsubsection{Densely Residual Feature Extraction}
\begin{figure}[htbp]
    \centering
    \includegraphics[width=3.5in,height=1.8in]{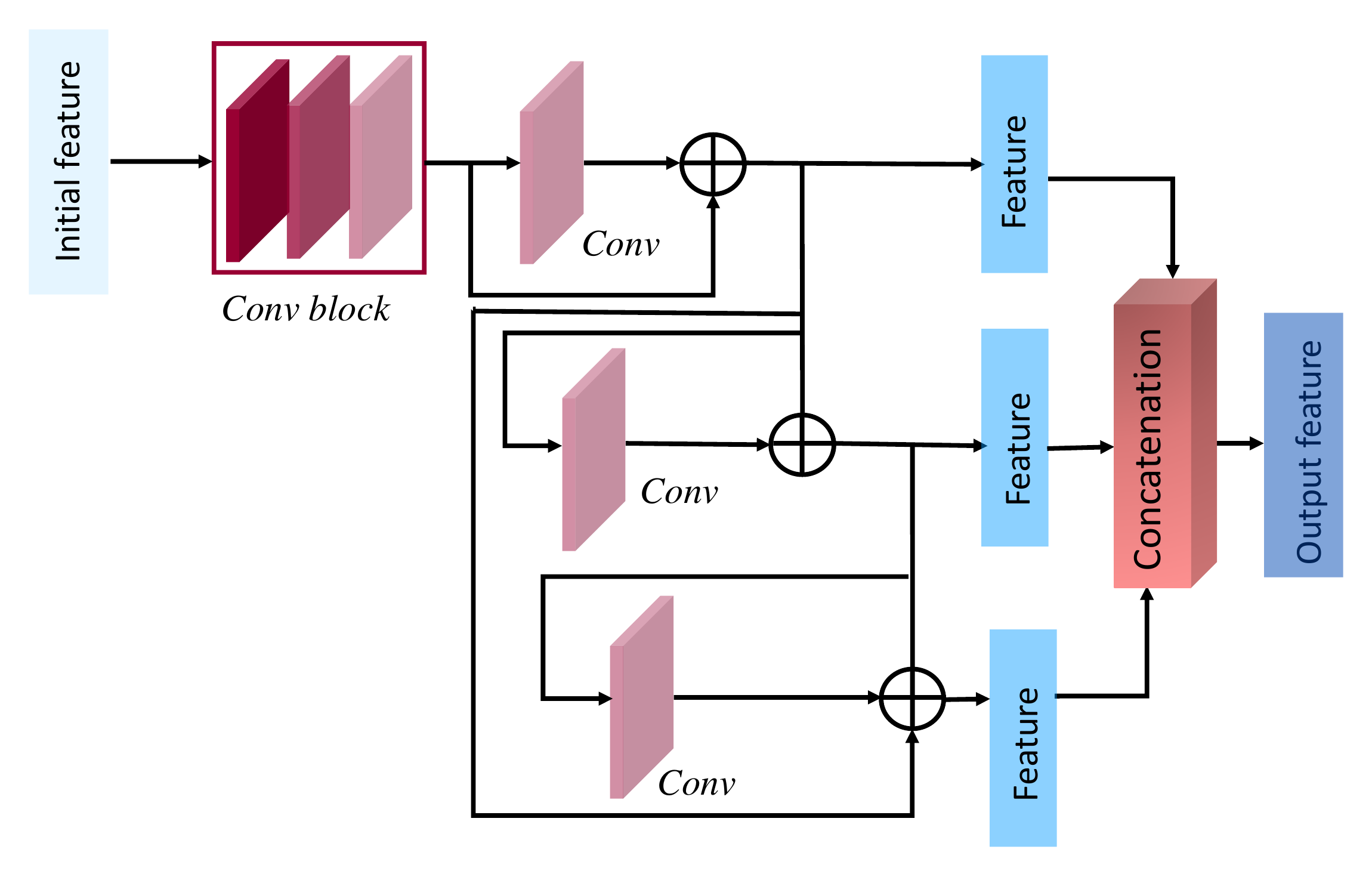}
    \caption{Densely residual feature extraction module}
    \label{densly_residual}
\end{figure}

The densely residual operation has shown great promise in both regression and classification tasks \cite{residual}. Dense residual connections are an efficient way to emphasize hierarchical representation. For this reason, we design a densely residual module to aggregate features for the network. The proposed design in fig. \ref{densly_residual} also utilizes the concatenation between the final and previous aggregation in support of total hierarchy concentration. A final convolution is added to combine the three concatenated features from the densely residual layers.

After collecting and concatenating the individual responses from each of the four modules, the responses are merged by the final convolution layer with a dilation rate of $2$, see the overall process in fig. \ref{architecture_overview}. This layer's output contains the most refined representation for the restored image. The restored image is fed into a simple loss function consisting of $\ell_1$ and $\ell_{SSIM}$.

\subsection{Loss Function}
We use two typical loss functions $\ell_1$ and $\ell_{SSIM}$ to update the parameter space. The total loss function is a simple addition of the two. 
\begin{equation}
    \ell_{total} = \ell_{1} + \ell_{SSIM}.
\end{equation}
 $\ell_1$ measures the distance between the ground truth clean image and the restored image as shown in next equation.

\begin{equation}
     \ell_{1} = \frac{1}{n}\sum_{j=1}^n|\gamma_{g}-\gamma_{p}|.
 \end{equation}
Here, $\gamma_{g}$ is the ground truth clean image and $\gamma_{p}$ is the restored prediction image. The secondary component is the loss function from SSIM, which is another widely used similarity measure for images. 
\begin{equation}
    \ell_{SSIM} = \frac{1}{n}\sum_{j=1}^n 1-SSIM(\gamma_g,\gamma_p)
\end{equation}

\section{Experimental Results}
This section describes the overall performance of our method on both real and synthetic noisy images. 


\begin{figure*}[!htbp]
\centering
\begin{subfigure}[l]{0.10\textwidth}
    \centering
    \includegraphics[height=18mm,width=18mm]{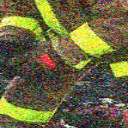}
    \caption{Fireman}
\end{subfigure}
\begin{subfigure}[r]{0.70\textwidth}
    \centering
    \begin{subfigure}[t]{0.13\textwidth}
        \centering
        \includegraphics[height=16mm,width=\textwidth]{figure_6/fireman_04.png}
        \subcaption{Noisy}
    \end{subfigure}
        \begin{subfigure}[t]{0.13\textwidth}
        \centering
        \includegraphics[height=16mm,width=\textwidth]{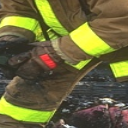}\\
        \caption{GT}
    \end{subfigure}
     \begin{subfigure}[t]{0.13\textwidth}
        \centering
        \includegraphics[height=16mm,width=\textwidth]{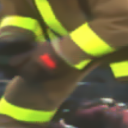}\\
        \caption{BM3D\cite{BM3D}}
    \end{subfigure}
     \begin{subfigure}[t]{0.13\textwidth}
        \centering
        \includegraphics[height=16mm,width=\textwidth]{figure_6/BM3D_patch1.png}\\
       \caption{WNNM\cite{WNNM}}
    \end{subfigure}
    \begin{subfigure}[t]{0.13\textwidth}
        \centering
        \includegraphics[height=16mm,width=\textwidth]{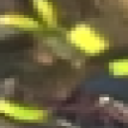}\\
       \caption{DnCNN\cite{DnCNN}}
    \end{subfigure}
    \begin{subfigure}[t]{0.13\textwidth}
        \centering
        \includegraphics[height=16mm,width=\textwidth]{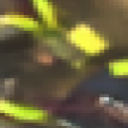}\\
        \caption{FFDNet\cite{ffdnet}}
    \end{subfigure}\\
        \begin{subfigure}[t]{0.13\textwidth}
        \centering
        \includegraphics[height=16mm,width=\textwidth]{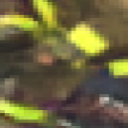}
        \caption{IrCNN\cite{IrCNN}}
    \end{subfigure}
        \begin{subfigure}[t]{0.13\textwidth}
        \centering
        \includegraphics[height=16mm,width=\textwidth]{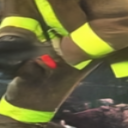}
       \caption{VDNet\cite{VDNet}}
    \end{subfigure}
        \begin{subfigure}[t]{0.13\textwidth}
        \centering
        \includegraphics[height=16mm,width=\textwidth]{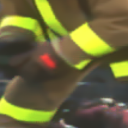}
        \caption{ADNet\cite{ADnet}}
    \end{subfigure}
    \begin{subfigure}[t]{0.13\textwidth}
        \centering
        \includegraphics[height=16mm,width=\textwidth]{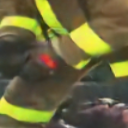}
        \caption{RIDNet\cite{RIDnet}}
    \end{subfigure}
    \begin{subfigure}[t]{0.13\textwidth}
        \centering
        \includegraphics[height=16mm,width=\textwidth]{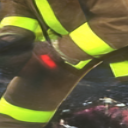}
       \caption{DeamNet\cite{deamnet}}
    \end{subfigure}
        \begin{subfigure}[t]{0.13\textwidth}
        \centering
        \includegraphics[height=16mm,width=\textwidth]{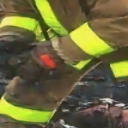}
       \caption{Proposed}
    \end{subfigure}
\end{subfigure}
\caption{Visual quality comparison for ``Fireman" from the BSD68 dataset with AWGN noise level $\sigma = 50$ (for best view, zoom-in is recommended).}
\label{bsd68}
\end{figure*}


\begin{figure*}[!htbp]
\centering
\begin{subfigure}[l]{0.10\textwidth}
    \centering
    \includegraphics[height=18mm,width=18mm]{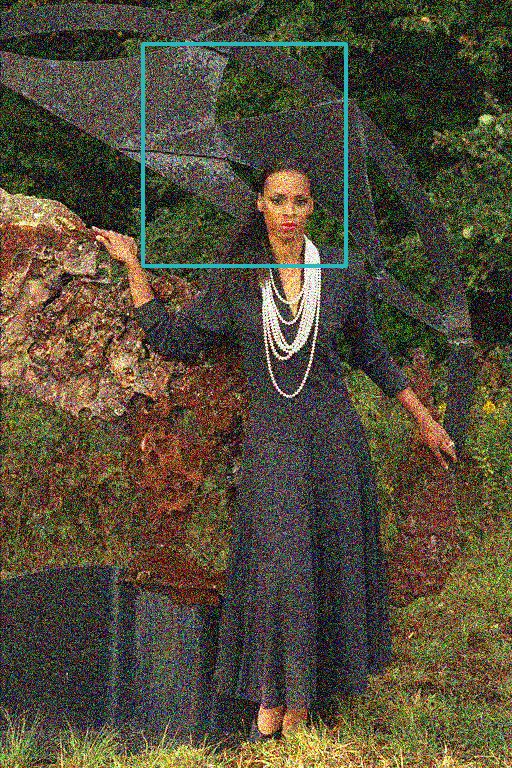}
    \caption{Model}
\end{subfigure}
\begin{subfigure}[r]{0.70\textwidth}
    \centering
    \begin{subfigure}[t]{0.13\textwidth}
        \centering
        \includegraphics[height=16mm,width=\textwidth]{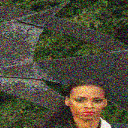}
        \subcaption{Noisy}
    \end{subfigure}
        \begin{subfigure}[t]{0.13\textwidth}
        \centering
        \includegraphics[height=16mm,width=\textwidth]{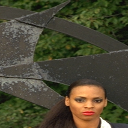}\\
        \caption{GT}
    \end{subfigure}
     \begin{subfigure}[t]{0.13\textwidth}
        \centering
        \includegraphics[height=16mm,width=\textwidth]{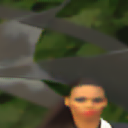}\\
        \caption{BM3D\cite{BM3D}}
    \end{subfigure}
     \begin{subfigure}[t]{0.13\textwidth}
        \centering
        \includegraphics[height=16mm,width=\textwidth]{figure_7/bm3d_kodak.png}\\
       \caption{WNNM\cite{WNNM}}
    \end{subfigure}
    \begin{subfigure}[t]{0.13\textwidth}
        \centering
        \includegraphics[height=16mm,width=\textwidth]{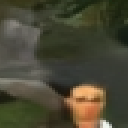}\\
       \caption{DnCNN\cite{DnCNN}}
    \end{subfigure}
    \begin{subfigure}[t]{0.13\textwidth}
        \centering
        \includegraphics[height=16mm,width=\textwidth]{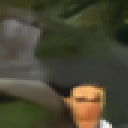}\\
        \caption{FFDNet\cite{ffdnet}}
    \end{subfigure}\\
        \begin{subfigure}[t]{0.13\textwidth}
        \centering
        \includegraphics[height=16mm,width=\textwidth]{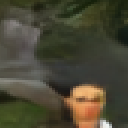}
        \caption{IrCNN\cite{IrCNN}}
    \end{subfigure}
        \begin{subfigure}[t]{0.13\textwidth}
        \centering
        \includegraphics[height=16mm,width=\textwidth]{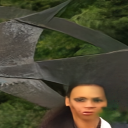}
       \caption{VDNet\cite{VDNet}}
    \end{subfigure}
        \begin{subfigure}[t]{0.13\textwidth}
        \centering
        \includegraphics[height=16mm,width=\textwidth]{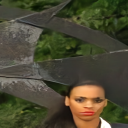}
        \caption{ADNet\cite{ADnet}}
    \end{subfigure}
    \begin{subfigure}[t]{0.13\textwidth}
        \centering
        \includegraphics[height=16mm,width=\textwidth]{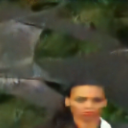}
        \caption{RIDNet\cite{RIDnet}}
    \end{subfigure}
    \begin{subfigure}[t]{0.13\textwidth}
        \centering
        \includegraphics[height=16mm,width=\textwidth]{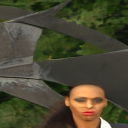}
       \caption{DeamNet\cite{deamnet}}
    \end{subfigure}
        \begin{subfigure}[t]{0.13\textwidth}
        \centering
        \includegraphics[height=16mm,width=\textwidth]{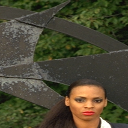}
       \caption{Proposed}
    \end{subfigure}
\end{subfigure}
\caption{Visual quality comparison for ``Model in black dress" from the Kodak24 dataset with AWGN noise level $\sigma = 50$ (for best view, zoom-in is recommended)..}
\label{kodak}
\end{figure*}

\subsection{Network Implementation and Training Set}
For the proposed study, we utilized a TensorFlow framework with NVIDIA GPU support. Most of the convolutional layers in our network are $3\times3$ kernels, apart from the specific cases where $1\times1$, $5\times5$, and $7\times7$ kernels in addition to the $3\times3$ kernels were used. For the training phase, we used He \textit{et. al} \cite{initialization} for initialization and the Adam optimizer with the learning rate $10^{-4}$, a typical default in many vision studies.

For the training, the DIV2K dataset was used. To enable diversity in the data flow, the typical rotation, blurring, contrast stretching, and inverse augmentation techniques were implemented. The training images were cropped into smaller patches. The noisy input images were created by perturbing the clean patches by AWGN noise with $15$, $25$, and $50$ standard deviations.

\subsection{Testing set}
We use the BSD68, Kodak24, and Urban100 datasets for the inference comparison, where clean observations are available and noisy versions are created through the same artificial noise augmentation. The results are summarized in table \ref{comparison table}.

The DND, SIDD, and RN15 datasets are used to evaluate the proposed approach on images with natural noise. A brief description of the real-world noisy image dataset and the evaluation procedures are described below.

\begin{itemize}
\item \textbf{DND}: DND\cite{DND} is a real-world image dataset consisting of $50$ real-world noisy images. However, near noise-free counterparts are unavailable to the public. The corresponding server provides the PSNR/SSIM results for the uploaded denoised images. 
\item \textbf{SIDD}: SIDD\cite{SIDD} is another real-world noisy image dataset that provides $320$ pairs of noisy images and near noise-free counterparts for training. This dataset follows similar evaluation process as DND dataset.
\item \textbf{RN15}: 
RN15\cite{NC} dataset provides $15$ real-world noisy images. Due to the unavailability of the ground truths, we only present the visual result of this dataset.
\end{itemize}


\begin{figure*}[!htbp]
\centering
\begin{subfigure}[l]{0.10\textwidth}
    \centering
    \includegraphics[height=18mm,width=18mm]{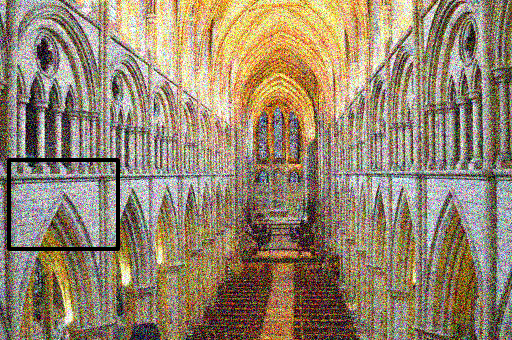}
    \caption{Interior}
\end{subfigure}
\begin{subfigure}[r]{0.70\textwidth}
    \centering
    \begin{subfigure}[t]{0.13\textwidth}
        \centering
        \includegraphics[height=16mm,width=\textwidth]{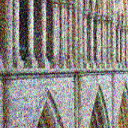}
        \subcaption{Noisy}
    \end{subfigure}
        \begin{subfigure}[t]{0.13\textwidth}
        \centering
        \includegraphics[height=16mm,width=\textwidth]{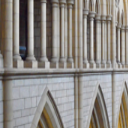}\\
        \caption{GT}
    \end{subfigure}
     \begin{subfigure}[t]{0.13\textwidth}
        \centering
        \includegraphics[height=16mm,width=\textwidth]{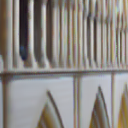}\\
        \caption{BM3D\cite{BM3D}}
    \end{subfigure}
     \begin{subfigure}[t]{0.13\textwidth}
        \centering
        \includegraphics[height=16mm,width=\textwidth]{figure_8/bm3d_urban.png}\\
       \caption{WNNM\cite{WNNM}}
    \end{subfigure}
    \begin{subfigure}[t]{0.13\textwidth}
        \centering
        \includegraphics[height=16mm,width=\textwidth]{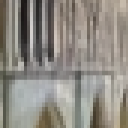}\\
       \caption{DnCNN\cite{DnCNN}}
    \end{subfigure}
    \begin{subfigure}[t]{0.13\textwidth}
        \centering
        \includegraphics[height=16mm,width=\textwidth]{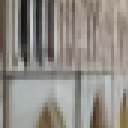}\\
        \caption{FFDNet\cite{ffdnet}}
    \end{subfigure}\\
        \begin{subfigure}[t]{0.13\textwidth}
        \centering
        \includegraphics[height=16mm,width=\textwidth]{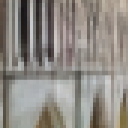}
        \caption{IrCNN\cite{IrCNN}}
    \end{subfigure}
        \begin{subfigure}[t]{0.13\textwidth}
        \centering
        \includegraphics[height=16mm,width=\textwidth]{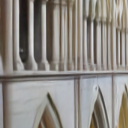}
       \caption{VDNet\cite{VDNet}}
    \end{subfigure}
        \begin{subfigure}[t]{0.13\textwidth}
        \centering
        \includegraphics[height=16mm,width=\textwidth]{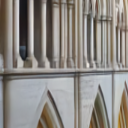}
        \caption{ADNet\cite{ADnet}}
    \end{subfigure}
    \begin{subfigure}[t]{0.13\textwidth}
        \centering
        \includegraphics[height=16mm,width=\textwidth]{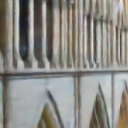}
        \caption{RIDNet\cite{RIDnet}}
    \end{subfigure}
    \begin{subfigure}[t]{0.13\textwidth}
        \centering
        \includegraphics[height=16mm,width=\textwidth]{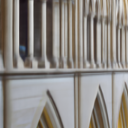}
       \caption{DeamNet\cite{deamnet}}
    \end{subfigure}
        \begin{subfigure}[t]{0.13\textwidth}
        \centering
        \includegraphics[height=16mm,width=\textwidth]{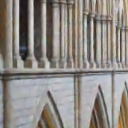}
       \caption{Proposed}
    \end{subfigure}
\end{subfigure}
\caption{Visual quality comparison for ``Interior" from the Urban100 dataset with AWGN noise level $\sigma = 50$ (for best view, zoom-in is recommended)..}
\label{urban100}
\end{figure*}

\subsection{Denoising on synthetic noisy images}
For evaluation purposes, we have considered previous state-of-the-art studies within various contexts. The evaluation procedure includes two filtering methods BM3D\cite{BM3D}, WNNM\cite{WNNM} and several convolutional networks including DnCNN\cite{DnCNN}, FFDNet\cite{ffdnet}, IrCNN\cite{IrCNN}, ADNet\cite{ADnet}, RIDNet\cite{RIDnet}, VDN\cite{VDNet}, and DEAMNet\cite{deamnet}.

Table \ref{comparison table} shows the average PSNR/SSIM scores for the quantitative comparison. From the average PSNR and SSIM score, the proposed study surpasses the previous studies with a considerable margin. We have adopted three widely used datasets BSD68, Kodak24, and Urban100 with three different AWGN noise levels $15$, $50$, and $50$.

For visual comparison, \ref{bsd68}, \ref{kodak}, and \ref{urban100} from BSD68, Kodak24, and Urban100 are presented respectively with the noise level of $50$. Figure \ref{bsd68} shows the ``fireman" picture from the BSD68 dataset. The differences in the restoration are shown in detail with more controlled smoothing. From figure \ref{kodak}, we see that the proposed approach avoids image cartoonization and preserves details while restoring clean details. The proposed study manages to restore the structural continuity compared to other methods while preserving appropriate color and contrast of the image. The last visual comparison for the synthetic noisy image is the ``Interior" picture from the Urban100 dataset, shown in fig. \ref{urban100}. For better illustration of differences, a zoomed image of the interior wall of the place is shown, where the proposed method manages to preserve the brick separating lines more clearly.

\subsection{Denoising on real-world noisy images}
The results for real-world noisy image restoration are presented in Table \ref{raw-comparison table}. Natural noise removal is challenging because the convoluted noises are not signal independent and vary within the spatial neighborhood.

\begin{figure*}[!htbp]
\centering
\begin{subfigure}[r]{0.70\textwidth}
    \centering
    \begin{subfigure}[t]{0.15\textwidth}
        \centering
        \includegraphics[height=16mm,width=\textwidth]{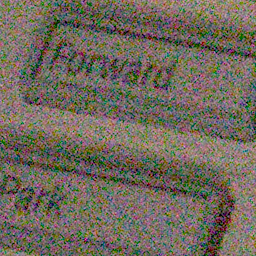}
        \subcaption{Noisy}
    \end{subfigure}
        \begin{subfigure}[t]{0.15\textwidth}
        \centering
        \includegraphics[height=16mm,width=\textwidth]{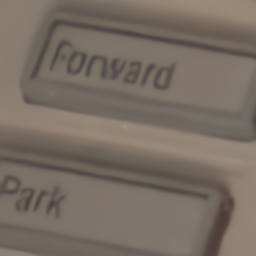}\\
        \caption{VDN\cite{VDNet}}
    \end{subfigure}
     \begin{subfigure}[t]{0.15\textwidth}
        \centering
        \includegraphics[height=16mm,width=\textwidth]{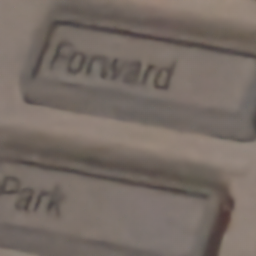}\\
        \caption{BM3D\cite{BM3D}}
    \end{subfigure}
     \begin{subfigure}[t]{0.15\textwidth}
        \centering
        \includegraphics[height=16mm,width=\textwidth]{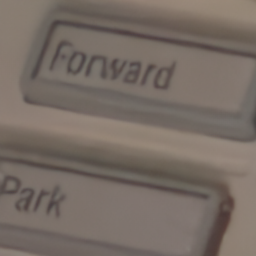}\\
       \caption{DEAMNet\cite{deamnet}}
    \end{subfigure}
    \begin{subfigure}[t]{0.15\textwidth}
        \centering
        \includegraphics[height=16mm,width=\textwidth]{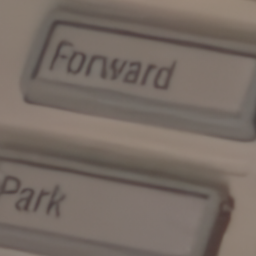}\\
       \caption{Proposed}
    \end{subfigure}
    
\end{subfigure}
\caption{Visual quality comparison for the SIDD dataset with real noises (for best view, zoom-in is recommended).}
\label{SIDD}
\end{figure*}



\begin{figure*}[!htbp]
\centering
\begin{subfigure}[l]{0.05\textwidth}
    \centering
    \includegraphics[height=20mm,width=20mm]{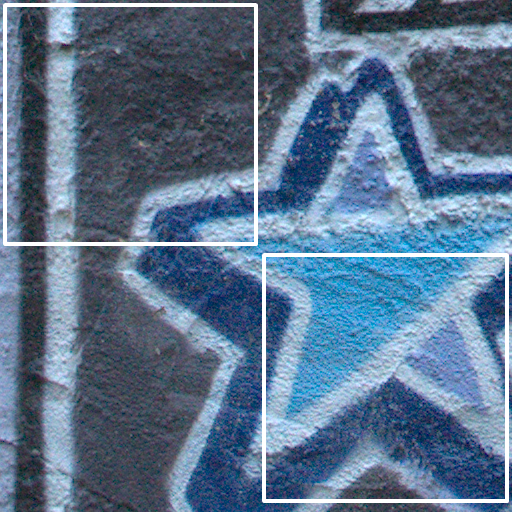}
    \subcaption{Star}
   
    \centering
\end{subfigure}
\begin{subfigure}[r]{0.70\textwidth}
    \centering
    \begin{subfigure}[t]{0.13\textwidth}
        \centering
        \includegraphics[height=16mm,width=\textwidth]{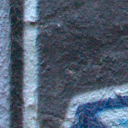}
      
    \end{subfigure}
        \begin{subfigure}[t]{0.13\textwidth}
        \centering
        \includegraphics[height=16mm,width=\textwidth]{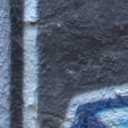}\\
      
    \end{subfigure}
     \begin{subfigure}[t]{0.13\textwidth}
        \centering
        \includegraphics[height=16mm,width=\textwidth]{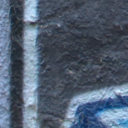}\\
        
    \end{subfigure}
     \begin{subfigure}[t]{0.13\textwidth}
        \centering
        \includegraphics[height=16mm,width=\textwidth]{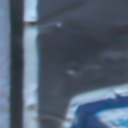}\\
       
    \end{subfigure}
    \begin{subfigure}[t]{0.13\textwidth}
        \centering
        \includegraphics[height=16mm,width=\textwidth]{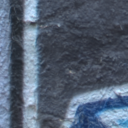}\\
        
    \end{subfigure}\\
    \begin{subfigure}[t]{0.13\textwidth}
        \centering
        \includegraphics[height=16mm,width=\textwidth]{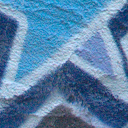}\\
        \caption{Noisy}
       
    \end{subfigure}
        \begin{subfigure}[t]{0.13\textwidth}
        \centering
        \includegraphics[height=16mm,width=\textwidth]{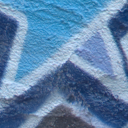}
        \caption{VDN\cite{VDNet}}
        
    \end{subfigure}
        \begin{subfigure}[t]{0.13\textwidth}
        \centering
        \includegraphics[height=16mm,width=\textwidth]{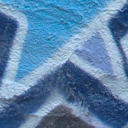}
       \caption{RIDNet\cite{RIDnet}}
      
    \end{subfigure}
        \begin{subfigure}[t]{0.13\textwidth}
        \centering
        \includegraphics[height=16mm,width=\textwidth]{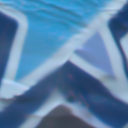}
        \caption{DeamNet\cite{deamnet}}
   
    \end{subfigure}
    \begin{subfigure}[t]{0.13\textwidth}
        \centering
        \includegraphics[height=16mm,width=\textwidth]{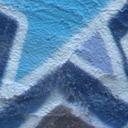}
        \caption{Proposed}
      
    \end{subfigure}
\end{subfigure}
\caption{Visual quality comparison for ``Star'' from the DnD dataset with real noises (for best view, zoom-in is recommended)}
\label{DnD}
\end{figure*}

We have chosen three real noisy image datasets, SIDD benchmark \cite{SIDD}, DnD benchmark\cite{DND}, and RN15 \cite{NC}, to analyze the generalization capability of our proposed method.  For SIDD and DnD benchmarks, the clean counterpart images are not openly distributed. Hence, the presented PSNR/SSIM in table \ref{raw-comparison table} is obtained by uploading the results into the corresponding server. For the RN15 dataset, there are no benchmark utility. Table \ref{raw-comparison table} represents comparative performance for both SIDD and DnD benchmark. Among the existing methods, VDN\cite{VDNet} and DEAMNet\cite{deamnet} performs well. However, our method achieves better result among the existing methods for both the real and synthetic noises.


\begin{figure}
     \centering
     \begin{subfigure}[t]{0.10\textwidth}
         \centering
         \includegraphics[width=\textwidth]{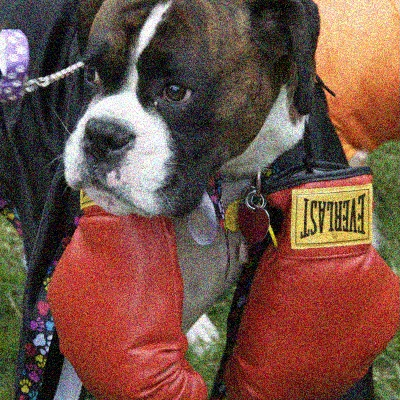}
         
     \end{subfigure}
     \begin{subfigure}[t]{0.10\textwidth}
         \centering
         \includegraphics[width=\textwidth]{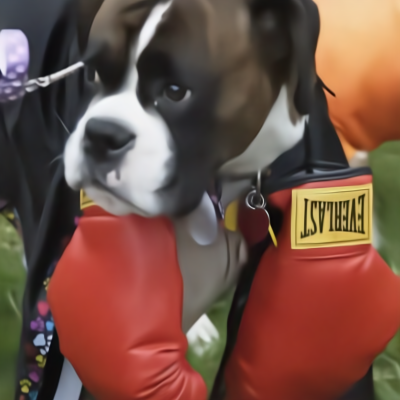}
         
     \end{subfigure}
     \begin{subfigure}[t]{0.10\textwidth}
         \centering
         \includegraphics[width=\textwidth]{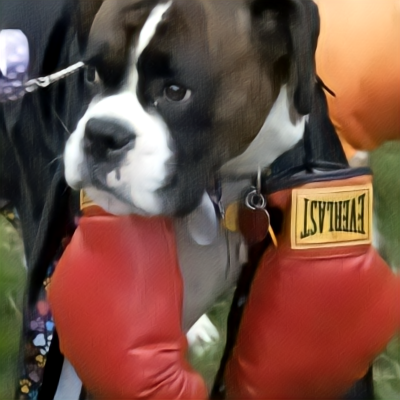}
        
     \end{subfigure}
     \begin{subfigure}[t]{0.10\textwidth}
         \centering
         \includegraphics[width=\textwidth]{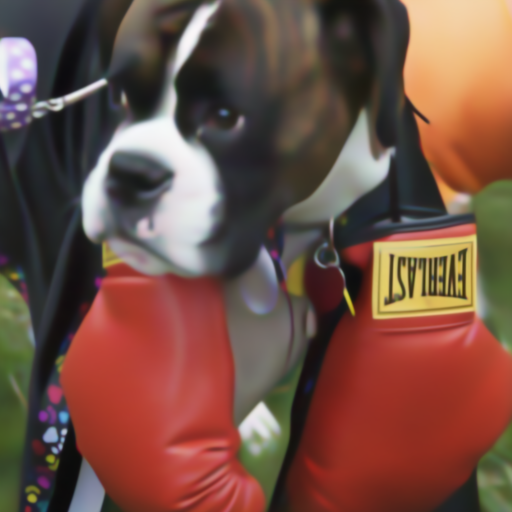}
         
     \end{subfigure}
     \begin{subfigure}[t]{0.10\textwidth}
         \centering
         \includegraphics[width=\textwidth]{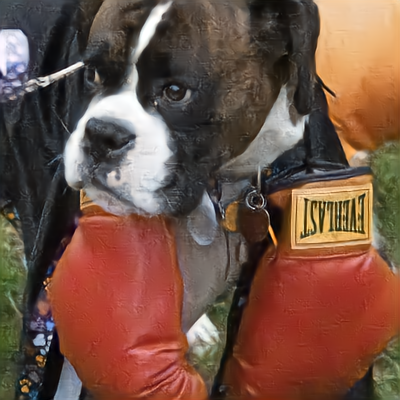}
         
     \end{subfigure}\\
     \begin{subfigure}[t]{0.10\textwidth}
         \centering
         \includegraphics[width=\textwidth]{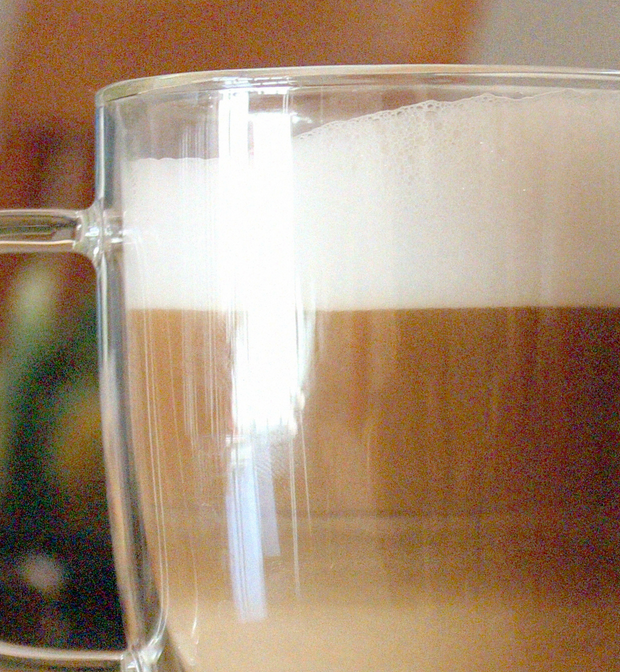}
         \subcaption{Noisy}
         
     \end{subfigure}
     \begin{subfigure}[t]{0.10\textwidth}
         \centering
         \includegraphics[width=\textwidth]{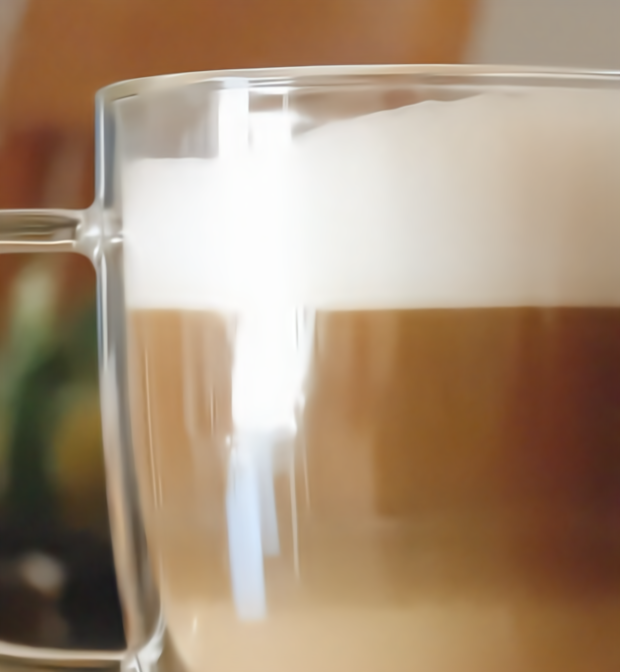}
         \subcaption{VDN\cite{VDNet}}
     \end{subfigure}
     \begin{subfigure}[t]{0.10\textwidth}
         \centering
         \includegraphics[width=\textwidth]{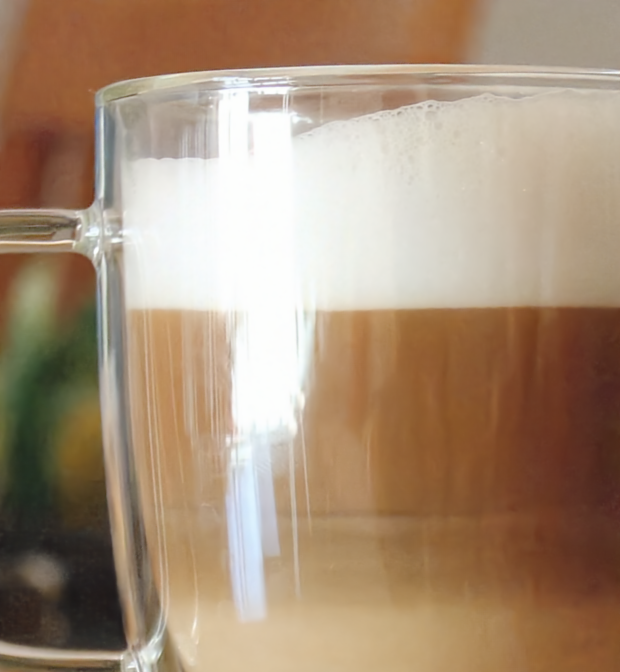}
         \subcaption{RIDNet\cite{RIDnet}}
         
     \end{subfigure}
     \begin{subfigure}[t]{0.10\textwidth}
         \centering
         \includegraphics[width=\textwidth]{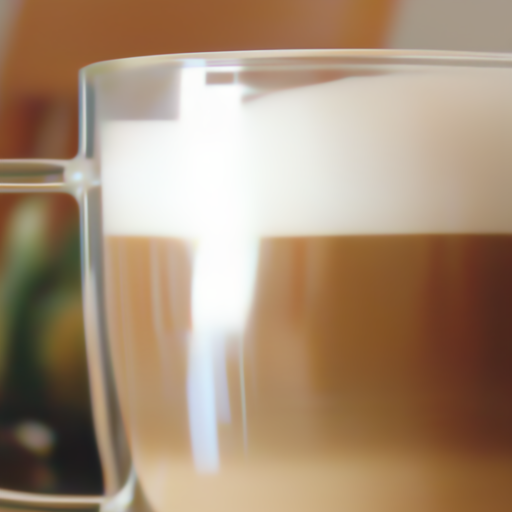}
         \subcaption{DeamNet\cite{deamnet}}
         
     \end{subfigure}
     \begin{subfigure}[t]{0.10\textwidth}
         \centering
         \includegraphics[width=\textwidth]{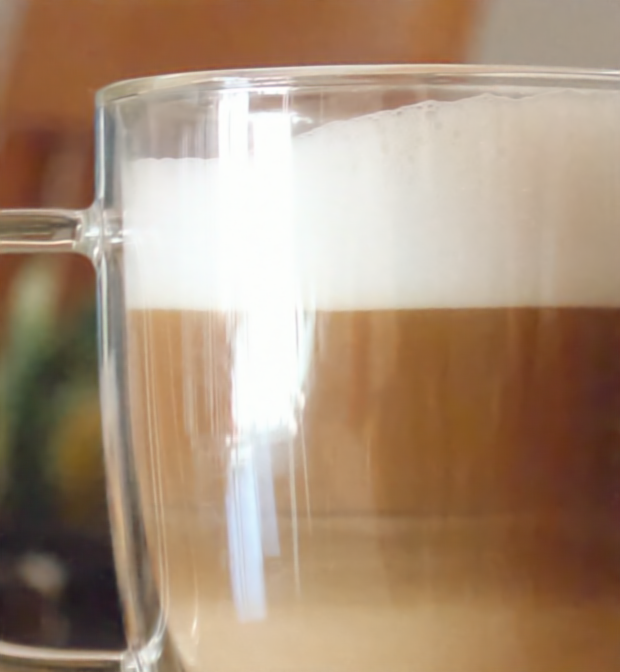}
         \subcaption{Proposed}
         
     \end{subfigure}\\
     
        \caption{Visual quality comparison for ``Dog"  from the RN15 dataset. RN15 dataset is set of real noise images without the clean image counterparts (for best view, zoom-in is recommended).}
        \label{rn15}
\end{figure}

To demonstrate our performance with real images, we have also provided some visual comparisons in figures \ref{SIDD}, \ref{DnD}, and \ref{rn15} on the SIDD, DnD, and RN15 datasets, respectively. For visual comparison on real noisy images, we have included the recent VDN\cite{VDNet}, RIDNet\cite{RIDnet}, and DEAMNet\cite{deamnet}. The visual comparison shows that our method tends to avoid cartoonization while effectively removing noise, suppressing artifacts, and preserving object edges. Overall, the qualitative and quantitative comparison displays effective performance in all fronts.


\begin{table*}[htbp]
\centering
\caption{Quantitative comparison results of the competing methods with AWGN noise level $\sigma =$15, 25, 50 on kodak24, BSD68, and Urban100. Top results are in bold, and second best result are underlined.}
\label{comparison table}
\resizebox{0.8\textwidth}{!}{%
\begin{tabular}{c|c|ccc|ccc|ccc}
    \hline
    \multirow{2}{*}{Method}&Metrics&&$\sigma =$ 15&&&$\sigma =$ 25&&&$\sigma =$ 50\\
    &&BSD68&Kodak24&Urban100&BSD68&Kodak24&Urban100&BSD68&Kodak24&Urban100\\
    \hline
    \multirow{2}{*}{BM3D\cite{BM3D}}&PSNR&32.37&31.07&32.35&29.97&28.57&29.70&26.72&25.62&25.95\\
                                    &SSIM&0.8952&0.8717&0.9220&0.8504&0.8013&0.8777&0.7676&0.6864&0.7791\\\hline
     \multirow{2}{*}{WNNM\cite{WNNM}}&PSNR&32.70&31.37&32.97&30.28&28.83&30.39&27.05&25.87&26.83\\
                                      &SSIM&0.8982&0.8766&0.9271&0.8577&0.8087&0.8885&0.7775&0.6982&0.8047\\\hline  
    
    \multirow{2}{*}{DnCNN\cite{DnCNN}}&PSNR&32.86&31.73&31.86&30.06&28.92&29.25&27.18&26.23&26.28\\
                                      &SSIM&0.9031&0.8907&0.9255&0.8622&0.8278&0.8797&0.7829&0.7189&0.7874\\\hline
    \multirow{2}{*}{FFDNet\cite{ffdnet}}&PSNR&32.75&31.63&32.43&30.43&29.19&29.92&27.32&26.29&26.28\\
    &SSIM&0.9027&0.8902&0.9273&0.8634&0.8289&0.8886&0.7903&0.7245&0.8057\\\hline
     \multirow{2}{*}{IrCNN\cite{IrCNN}}&PSNR&32.67&33.60&31.85  &29.96&\underline{30.98}&28.92   &26.59&27.66&25.21\\
                          &SSIM &0.9318&0.9247&\underline{0.9493}   &\underline{0.8859}&0.8799&0.9101   &0.7899&\underline{0.7914}&0.8168\\\hline

    \multirow{2}{*}{ADNet\cite{ADnet}}&PSNR&32.98&31.74&32.87&30.58&29.25&30.24&27.37&26.29&26.64\\
    &SSIM&0.9050&0.8916&0.9308&0.8654&0.8294&0.8923&0.7908&0.7216&0.8073\\\hline
    
    \multirow{2}{*}{RIDNet\cite{RIDnet}}&PSNR&32.91&31.81&33.11&30.60&29.34&30.49&27.43&26.40&26.73\\
    &SSIM&0.9059&0.8934&0.9339&0.8672&0.8331&0.8975&0.7932&0.7267&0.8132\\\hline
    \multirow{2}{*}{VDN\cite{VDNet}}&PSNR&\textbf{33.90}&\textbf{34.81}&\underline{33.41}&\underline{31.25}&\textbf{32.38}&30.83&\underline{28.19}&\textbf{29.19}&\textbf{28.43}\\
                                   &SSIM&\underline{0.9243}&\underline{0.9251}&0.9339&0.8713&\underline{0.8842}&0.8361&\underline{0.8014}&0.7213&0.8212\\\hline
    \multirow{2}{*}{DEAMNet\cite{deamnet}}&PSNR&33.19&31.91&33.37&30.81&29.44&\underline{30.85}&27.74&26.54&27.53\\
    &SSIM&0.9097&0.8957&0.9372&0.8717&0.8373&\underline{0.9048}&0.8057&0.7368&\underline{0.8373}\\\hline
    \multirow{2}{*}{Proposed}&PSNR&\underline{33.85}&\underline{32.90}&\textbf{33.97}      &\textbf{31.32}&30.67&\textbf{31.52}    &\textbf{29.02}&\underline{28.12}&\underline{28.25}\\
                            &SSIM&\textbf{0.9603}&\textbf{0.9517}&\textbf{0.9621 }   &\textbf{0.9150}&\textbf{0.9246}&\textbf{0.9241} &\textbf{0.8831}&\textbf{0.8782}&\textbf{0.8755}\\

    \hline
\end{tabular}
\label{comparison table1}
}

\end{table*}

\begin{table}[htbp]
\centering
\caption{Real-image denoising results of several existing methods on SIDD and DnD dataset. Top results are in bold, and second best results are underlined.}
\resizebox{0.8\columnwidth}{!}{%
\begin{tabular}{c|c|cccccccc}
    \hline
    Method&Metrics&BM3D&DnCNN&FFDNet&VDN&RIDNet&DEAMNet&Proposed\\
    \hline
    \multirow{2}{*}{SIDD\cite{SIDD}}&PSNR&25.65&23.66&29.30&39.26&37.87&\underline{39.35}&
    \textbf{39.55}\\
                         &SSIM&0.685&0.583&0.694&0.944&0.943&\underline{0.955}&\textbf{0.964}\\

    \hline
     \multirow{2}{*}{DnD\cite{DND}}&PSNR&34.51&32.43&37.61&39.38&39.25&\underline{39.63}&\textbf{39.76}\\
                         &SSIM&0.8507&0.7900&0.9115&0.9518&0.9528&\underline{0.9531}&\textbf{0.9617}\\

    \hline
\end{tabular}
\label{raw-comparison table}
}

\end{table}


\begin{table}[htbp]

    \centering
    \caption{Running time (in seconds) and parameter comparison}
    \resizebox{0.5\columnwidth}{!}{%
    \begin{tabular}{ccccc}
    \hline
    Method            &  Size $256^2$  &  Size $512^2$  & Size $1024^2$ & parameters\\
    \hline
    BM3D\cite{BM3D}   & 0.76           & 3.12           &12.82          &   -       \\
    WNNM\cite{WNNM}   & 210.26          & 858.04          &3603.68          &   -       \\
    DnCNN\cite{DnCNN} & 0.01           & 0.05           &0.16           &  558k     \\
    IrCNN\cite{IrCNN} & 0.012          & 0.038          &0.146          &   -       \\
    FFDNet\cite{ffdnet} & 0.01         & 0.05           &0.11           & 490k      \\
    AINDNet\cite{aindnet}&0.05         & 0.03           &0.80           & 13764k    \\
    ADNet\cite{ADnet} &0.02            &0.06            &0.20           & 519k      \\
    VDN\cite{VDNet}  & 0.04            &0.07            &0.19           & 7817k     \\
    RIDNet\cite{RIDnet}&0.07           & 0.21           &0.84           & 1499k     \\
    DEAMNet\cite{deamnet}&0.05         &0.19            &0.73           & 2225k     \\
    Proposed          &0.031           & 0.11           & 0.42          & 846k   \\
    \hline
    
    \end{tabular}
    }
    
    \label{running time }
    
\end{table}

\subsection{Computational Complexity}
This section provides a comparison of computational complexity through table \ref{running time }. The table represents the average running times for the three different image sizes $256\times256, 512\times512$, and $1024\times1024$. In addition, we present the parameter counts of the compared methods. Apart from BM3D\cite{BM3D}, we report model-specific computation time. In this comparison, we have considered, BM3D\cite{BM3D}, DnCNN\cite{DnCNN}, WNNM\cite{WNNM}, IrCNN\cite{IrCNN}, FFDNet\cite{ffdnet}, AINDNet\cite{aindnet}, ADNet\cite{ADnet}, VDN\cite{VDNet}, RIDNet\cite{RIDnet}, and DEAMNet\cite{deamnet}. In Table \ref{running time }, our method's computation time is only slightly longer than earlier DnCNN, IrCNN, FFDNet, ADNet and VDN. In terms of parameter counting, the proposed study is significantly smaller than the recent RIDNet\cite{RIDnet}, AINDNet\cite{aindnet}, VDN\cite{VDNet}, and DEAMNet\cite{deamnet}.

\section{Conclusion}
In this paper, the basic strategy for the lowlevel denoising problem is to gather a variety of lowlevel features while keeping the interpretation simple by implementing relatively shallow layers. We argue that for lowlevel vision tasks, the principle of Occam's razor is more appropriate, and accordingly, we have designed a network that focused on gathering a variety of lowlevel evidence rather than providing an deep explanation of the evidence. Thus, we have revisited the feature ensemble approach for the image denoising problem. Our study offers a new model which concatenates different modules for creating large and varying feature maps. To enhance the performance of our network, we utilized different kernel sizes, residual and densely-residual connections, and avoided deep unimodule cascaded aggregation. We carefully designed four different modules for our study, where each helps to restore different spatial properties. Finally, we have validated our network with natural and synthetic noisy images. Extensive comparisons show the overall efficiency of the proposed study.

\bibliographystyle{unsrt}  
\bibliography{references}

\end{document}